\documentclass[a4paper,11pt,final]{report}

\usepackage{subfig}
\usepackage[english]{babel}
\usepackage{indentfirst}
\usepackage[utf8]{inputenc}
\usepackage[T1]{fontenc}
\usepackage[pdftex]{graphicx}
\usepackage{setspace}
\usepackage{hyperref}
\usepackage[english]{varioref}
\usepackage{mathtools}
\usepackage{float}
\usepackage{newclude}
\usepackage{wrapfig}
\usepackage{setspace}
\usepackage{rotating}
\usepackage[table,xcdraw,dvipsnames]{xcolor}
\usepackage{caption}
\usepackage[toc,page]{appendix}
\usepackage[english]{minitoc}
\usepackage{verbatim}
\usepackage[top=3cm, bottom=3cm, left=3.2cm, right=3.2cm]{geometry}
\usepackage{fancyhdr}
\usepackage[Sonny]{fncychap}
\usepackage{ragged2e}

\usepackage{amsmath,amsfonts,amsthm} % Math packages

\usepackage{tabularx}

\usepackage[numbers]{natbib}

\usepackage{algorithm}
\usepackage{algpseudocode}

\usepackage{tikz}

\usepackage{multicol}
\usepackage{multirow}
\usepackage{booktabs}

\newcounter{ALC@tempcntr}% Temporary counter for storage

\newcolumntype{P}[1]{>{\raggedright\arraybackslash}p{#1}}

\usepackage{color,soul}

\definecolor{softgray}{RGB}{220,220,220}
\definecolor{color1}{RGB}{0,0,90} % Color of the article title and sections
\definecolor{color2}{RGB}{0,20,20} % Color of the boxes behind the abstract and headings

\usepackage{pdfpages}

\algnewcommand{\IIf}[1]{\State\algorithmicif\ #1\ \algorithmicthen}
\algnewcommand{\EndIIf}{\unskip\ \algorithmicend\ \algorithmicif}
\algnewcommand{\ElseI}{\ \algorithmicelse\ }

\algrenewcommand{\alglinenumber}[1]{\footnotesize#1}

\newcommand{\ignore}[1]{}

\newcommand{\HRule}{\rule{\linewidth}{0.5mm}}
\newenvironment{claim}[1]{\par\noindent\underline{Claim:}\space#1}{}

\theoremstyle{definition}
\newtheorem{definition}{Definition}[]

\begin{document}

% \dominitoc
\setcounter{mtc}{1}

% \includepdf{status-lua}
%\includepdf[pages=1]{cover-fr}

\centering{
  \noindent
  \HRule\\[0.4cm]
  \LARGE \textbf{\setstretch{1.0}
    A study of problems with multiple interdependent components --- Part I}\\[0cm]
  \LARGE
  \HRule \\[1cm]
  Mohamed \textsc{El Yafrani}
}\\[2cm]

\raggedright

\section*{Acknowledgement}

This document contains the first part of my Ph.D. dissertation.
This work have been prepared within the laboratory of research in computer science
and telecommunications (LRIT) at Mohammed V University under the supervision of
Pr. Belaïd \textsc{Ahiod}.
This thesis was defended on 14 September 2018, before the following board members:
\begin{itemize}
\item Abdelhakim Ameur \textsc{El Imrani}, Jury president, Professor, Mohammed V University in Rabat
\item Salma \textsc{Mouline}, Reporter, Professor, Mohammed V University in Rabat
\item Mohamed \textsc{Ouzineb}, Examiner, Habilitated professor, INSEA
\item Markus \textsc{Wagner}, Reporter, Senior Lecturer, The University of Adelaide
\item Myriam \textsc{Delgado}, Examiner, Professor, Federal University of Technology of Paraná
\item Belaïd \textsc{Ahiod}, Advisor, Habilitated professor, Mohammed V University in Rabat
\end{itemize}

\textbf{Contributors to part I:}\\
Belaïd \textsc{Ahiod}, Mohammed V University in Rabat\\
Mohammad Reza \textsc{Bonyadi}, Rio Tinto

\newpage

\pagenumbering{roman}
\setcounter{page}{1}

% \addcontentsline{toc}{chapter}{Acknowledgement}
%\include{acknowledgement}
%\newpage

% \addcontentsline{toc}{chapter}{Abstract}
\selectlanguage{english}

\begin{abstract}
\justify{
Recognising that real-world optimisation problems have multiple interdependent components
can be quite easy. However, providing a generic and formal model for dependencies between
components can be a tricky task. In fact, a PMIC can be considered simply as a single
optimisation problem and the dependencies between components could be investigated
by studying the decomposability of the problem and the correlations between the sub-problems.
In this work, we attempt to define PMICs by reasoning from a reverse perspective.
Instead of considering a decomposable problem, we model multiple problems (the components)
and define how these components could be connected.
In this document, we introduce notions related to problems with mutliple interndependent
components. We start by introducing realistic examples from logistics and supply chain
management to illustrate the composite nature and dependencies in these problems.
Afterwards, we provide our attempt to formalise and classify dependency in multi-component
problems.
}

\noindent \textbf{Keywords:} Interdependence, Multi-component problems, Combinatorial optimisation.

\end{abstract}

\newpage

% \addcontentsline{toc}{chapter}{Table des matières}
%\tableofcontents
% \mtcaddchapter
%\newpage

%--------------------------
% config header/footer
%--------------------------
 \newcommand{\brieftitle}{}
 \newcommand{\headertitle}{\brieftitle}
\pagestyle{fancy}
\fancyhf{}
\rhead{\thepage}
\lhead{\headertitle}

\pagenumbering{arabic}
\setcounter{page}{1}

%\include*{part0}

% PART I
%\part{Multi-component problems with dependencies}

\chapter{Multi-component problems with internal dependencies: illustrative examples and high-level definitions \label{chap:1}}

\renewcommand{\headertitle}{\nameref{chap:1}}

A problem with multiple interdependent components is an optimisation problem that embeds multiple sub-problems, usually called components, where the components can not be solved in isolation. This means that solving each component to optimality does not necessarily guarantee obtaining an optimal overall solution if the other components are not considered.

In this chapter, we provide an informal introduction using two illustrative examples. The examples are designed in order to reflect aspects from real-world optimisation problems, while kept as simple as possible.

%%===================================================
%% Illustrative example and informal definition
%%===================================================
% \section{Illustrative example and informal definition}

% the location routing problem as an example
\section{The problem of finding warehouse locations and distribution routes}

Let us start by considering the following problem inspired by the Location-Routing Problem~\cite{perl1985lrp}. Informally, the problem states as follows.

\textit{Given a set of $m$ potential distribution center (DC) locations and the associated cost for establishing a warehouse, a set of $n$ clients and their associated requests. Each client is to be associated to a regional DC. The size of the warehouse depends on the number of clients it serves. Therefore, the cost of establishing a warehouse also depends on the number of associated clients. Deliveries are done in routes where a vehicle serves multiple clients in one trip. We suppose that each DC has one vehicle to serve all the clients, and that the transportation cost depends on the total distances of the trips. }

The goal is to find a set of $k$ locations (where $k$ is a constant lower than $n$) and a set of routes where each client is served by its regional DC, such that the warehouse establishing and the transportation costs are minimised.

\subsection{Composition}
Given the above statement, it is clear that the problem described embeds the following two components.
\begin{itemize}
\item \textbf{A variant of the Facility Location Problem (FLP)~\cite{cornuejols1983uncapacitated}:} which corresponds to the problem of finding a set of $k$ locations to establish the warehouses such that it covers a regional area and reaches as many clients as possible. A solution for this problem is represented as a set of $k$ locations and thier associated clients.
\item \textbf{A variant of the Vehicle Routing Problem (VRP)~\cite{dantzig1959truck}:} which is the problem of finding a route (where a route is a set of ordered clients) for each DC such that each client gets served by a regional DC~\cite{renaud1996tabu}. A solution for this VRP variant is a set of $k$ routes.
\end{itemize}

The overall objective function can be expressed as the sum of the two components:

\textit{\textbf{Minimise} sum of distances between client and facility + facility establishment costs + sum of route distances}

\subsection{Dependencies between the components}
The interdependence between the two components can be easily recognised. 
In order to better understand the dependencies between the VRP and the FLP, let us consider the following scenarios.

\begin{itemize}
\item \textbf{Changing a warehouse location} might change the total distance of the routes. Meaning that changing the FLP solution might impact the objective value of the VRP solution.
\item \textbf{Changing a client's region} makes the routes associated non-feasible. Meaning that changing the FLP solution impacts the feasibility of the VRP solution.
\item \textbf{Changing the routes by assigning a client to another route} makes the set of associated clients non-feasible. This means that changing the VRP solution impacts the feasibility of the FLP solution.
\end{itemize}

What we can learn from the above problem is that in real-world situations, optimisation problems can be composed of multiple interdependent components. Furthermore, in the example we introduced two types of dependencies: the first one influences the objective value of solutions (\textit{fitness dependency}), while the second influences their feasibility (\textit{feasibility dependency}).

\section{The problem of demand scheduling and truck loading}

As a second illustrative example, let us consider the logistics problem expressed in the following statement.

\textit{A company owns a plant containing $m$ machines.
The plant manfactures on-demand items of $p$ products by processing every item on every machine given a specific order. The plant must follow a weekly schedule in order to manufacture its products. At the end of every week, the finished items (defined by their volumes) should be delivered to be loaded in rented containers in order to make deliveries. This process of manufacturing and containers loading is repeated every week until the overall demand is satisfied.}

The goal is to find a set of week schedules and a set of packing plans such that the delay and renting rate of containers are minimised.

\subsection{Composition}
It is clear that the described problem is composed of the following sub-problems.
\begin{itemize}
\item \textbf{The Job Shop Scheduling Problem (JSSP)~\cite{graham1966bounds}:} which corresponds to the problem of finding a week schedule that assigns manufacturing tasks to machines.
\item \textbf{The Bin Packing Problem (BPP)~\cite{johnson1973near}:} which corresponds to the problem of finding the optimal packing plans for containers.
\end{itemize}

In a sense, due to the time window constraint, the problem is composed of many JSSP and BPP sub-problems. In fact, the number of these sub-problems is in its own a variable and should be minimised, which translates to minimising the overall delay. Therefore, the objective function of the overall problem can be defined as follows:

\textit{\textbf{Minimise} total delay + containers renting rate}

\subsection{Dependencies between the components}

The concept of dependecy in this problem is very different from the one explained in the first example. In fact, changing a schedule of a given week may not only change the feasibility or objective value of the BPP solution; but it can change the dimension of the BPP problem.

Let us explain: when changing a week schedule --say to another one that produces more items-- the number of items to be loaded into containers increases, i.e., the problem dimension increases. We will refer to this type of dependency as \textit{Time dependency}.

The same conclusions can be drawn between the schedule for a given week and the Job Shop Scheduling Problem of the following week. Indeed, changing the schedule for a given week $w$ may change the number of tasks to be scheduled for the week $w+1$. In other words, the dimension of the JSSP is dependent on the schedule adopted the week before.

% \section{High-level definition}
Based on these two illustrative examples, we can informally define a problem with multiple interdependent component as a problem composed of multiple sub-problems that are non-separable. The non-separability can be due to three factors: impact on the objective value, impact on feasibility and impact on the dimension of a sub-problem.

\section{What differentiates multi-component problems from single-component ones?}

Problems with multiple interdependent components can be simply thought of as another way to look at decomposition and non-separability. However, there are some key features that distinguish problems with multiple interdependent components from previous related works:

\begin{enumerate}
\item \textbf{Intractability of components:} there are many existing examples allowing to argue that real-world problems are composed of sub-problems that are themselves intractable~\cite{perl1985lrp,iori2010routing,michalewicz2009applications,ibrahimov2012evolutionary1,ibrahimov2012evolutionary2,polyakovskiy2017just,stolk2013combining}. As mentioned earlier, we are interested in problems where all the components are $\mathcal{NP}$-complete. 
\item \textbf{Heterogeneity or natural decomposability of the overall problem:} meaning that components can be considered as different optimisation problems. While this may not be of interest from a computational complexity perspective (since all the components are $\mathcal{NP}$-complete), it is highly important from a modelling, representation and heuristic point-of-view. In fact, some $\mathcal{NP}$-complete problems are harder to model, represent and find approximate solutions for.
\end{enumerate}

\section{Conclusion}

The main goal of this chapter was to introduce the idea of dependency between component through two comprehensive examples.
In the next chapter, we provide our attempt to formally define the dependencies between components in an optimisation problem.

\chapter{A formal model for problems with multiple interdependent components\label{chap:2}}

\renewcommand{\headertitle}{\nameref{chap:2}}

% A problem with multiple interdependent components is an optimisation problem that embeds multiple subproblems, usually called components, where the components can not be solved in isolation. This means that solving each component to optimality does not necessarily guarantee obtaining an optimal overall solution if the other components are not considered.

In this chapter, we propose formal definitions of problems with multiple interdependent components and classify three forms of dependencies. Our model builds on definitions introduced by~\citet{bonyadi2016evolutionary}, while extending and formalising the concepts with the aim of presenting a more accurate model.

%%===================================================
%% Formal definitions
%%===================================================
\section{Preliminaries}

% In order to define the dependency relashionship, 

Let $P = \{P_1, \dots, P_n\}$ be a set of minimisation problems called \textit{components}, where a component $P_i, i \in \{1, \dots, n\}$, is defined by the following.

\begin{itemize}

% \item We suppose that the dimension of $P_i$, denoted $m^{P_i}$, is fixed.

\item We note $I^{P_i}$ the set of all possible $P_i$ instances with a fixed dimension denoted $m^{P_i}$.

\item $\mathcal{S}^{P_i}$ denotes the set of all possible solution configurations of $P_i$. That is, using an $N$-bitstring representation, the set of all $2^N$ possible configurations.

\item Given an instance $x \in I^{P_i}$, we note $F^{P_i}(x) \subseteq S^{P_i}$ the set of feasible solutions.

\item $\chi^{P_i}$ is a function that associates an instance from $I^{P_i}$ to a given $(n-1)$-tuple containing a solution configuration for every other component $P_j, j \in \{1, \dots, n\} \setminus \{i\}$ (Eq.~\ref{eq:chi-comp}).
  \begin{align}\label{eq:chi-comp}
    \chi^{P_1} \colon \mathcal{S}^{P_1}\times \dots \times \mathcal{S}^{P_{i-1}} \times \mathcal{S}^{P_{i+1}} \times \dots \times \mathcal{S}^{P_n} &\to I^{P_i}
  \end{align}

\item We note $Z^{P_i}$ the component objective function of $P_i$ 
  that associates a real number $z$ to an instance $x$ of $P_i$, a solution $s$ of $x$, 
  and $n-1$ solution configurations for every other problem $P_j, j \in \{1, \dots, n\}$, 
  as expressed in Equation~\ref{eq:obj1}.
  \begin{align}\label{eq:obj1}
    Z^{P_i} \colon 
    \chi^{P_i}(\overline{s}^{P_1}, \dots, \overline{s}^{P_{i-1}}, \overline{s}^{P_{i+1}}, \dots, \overline{s}^{P_n})
      \times\\\nonumber
    F^{P_i}(x) 
      \times\\\nonumber
    \mathcal{S}^{P_1}\times \dots \times \mathcal{S}^{P_{i-1}} \times \mathcal{S}^{P_{i+1}} \times \dots \times \mathcal{S}^{P_n}
    &\to \mathbb{R}\\\nonumber
    (x, s, \overline{s}^{P_1}, \dots, \overline{s}^{P_{i-1}}, \overline{s}^{P_{i+1}}, \dots, \overline{s}^{P_n})
      &\mapsto
    z\nonumber
  \end{align}
  
\end{itemize}

\section{Instance-dependency --- Dependency of fitness and feasibility}

We start by defining a general type of dependency where a change of solution for one problem impacts the instance of the other problem (Definition~\ref{def:instance dependency}).

\theoremstyle{definition}
\begin{definition}[Instance-dependency]\label{def:instance dependency}
Let $P_i$ and $P_j$ be two components in $P$.
We say that $P_i$ is instance-dependent on $P_j$ (notation $P_i \xleftarrow{}{} P_j$) if
$\exists s^{P_j}_1, s^{P_j}_2 \in \mathcal{S}^{P_j}$ such that:
\begin{align}
\chi^{P_i}(\overline{s}^{P_1},\dots, s^{P_j}_1, \dots, \overline{s}^{P_n})
\neq
\chi^{P_i}(\overline{s}^{P_1},\dots, s^{P_j}_2, \dots, \overline{s}^{P_n})
\nonumber
\end{align}
$\forall \overline{s}^{P_k} \in \mathcal{S}^{P_k}, \forall k \in \{1,\dots,n\} \setminus \{i,j\}$
\end{definition}

The first illustrative example in chapter~\ref{chap:1} demonstrated two forms of dependencies. The first is when a change of the solution on one problem impacts the fitnesses of solutions for the other problem as expressed in Definition~\ref{def:fitness-dependency}. While the second can be seen as a stronger form of instance dependency where the impact of changing the solution of a problem impact the feasibility of the solutions as shown in Definition~\ref{def:feasibility-dependency}. In our definitions, we conjecture that an instance dependency can only occur in one of these two forms.

\begin{definition}[Fitness-dependency]\label{def:fitness-dependency}
Let $P_i$ and $P_j$ be two components in $P$.
We say that $P_i$ is fitness-dependent on $P_j$ (notation $P_i \xleftarrow{fitness}{} P_j$) if
\begin{itemize}
\item $P_i \xleftarrow{}{} P_j$.
\item $\forall x_1, x_2 \in I^{P_i}, F^{P_i}(x_1)=F^{P_i}(x_2)$.
\end{itemize}
\end{definition}

\begin{definition}[Feasibility-dependency]\label{def:feasibility-dependency}
Let $P_i$ and $P_j$ be two components in $P$.
We say that $P_i$ is feasibility-dependent on $P_j$ (notation $P_i \xleftarrow{feasibility}{} P_j$) if
\begin{itemize}
\item $P_i \xleftarrow{}{} P_j$.
\item $\exists x_1, x_2 \in I^{P_i}, F^{P_i}(x_1) \neq F^{P_i}(x_2)$.
\end{itemize}
\end{definition}

Note that when both fitness- and feasibility-dependency are possible between two problems (for instance by considering two different change operators), fitness-dependency can be ommited as it is considered a weaker form of dependency.

Using Definitions \ref{def:fitness-dependency} and \ref{def:feasibility-dependency}, we can construct a directed graph $G = (P, D)$, called the graph of dependencies, consisting of the set of nodes $P$ which corresponds to the set of problems previously mentioned; and the set $D$ of edges, which are ordered pairs of elements of $P$, and which represent the dependencies between the problems.

If the graph $G$ is connected, we call $P$ a \textit{multi-component problem with internal dependencies} 
or a \textit{problem with multiple interdependent components}\footnote{the word interdependent here stands for internal dependencies and not for multual dependencies},
$P_i$ a component of $P$, and $Z^{P_i}$ a component objective function. 
We can then define either the overall objective function of $P$, noted $Z^{P}$ in terms of component objectives (Equation \ref{eq:overall-objective}); or consider $P$ as a multi-objective problem where the objective functions correspond to the component objective functions.

\begin{figure}[h]
  \centering
  \includegraphics[scale=0.5]{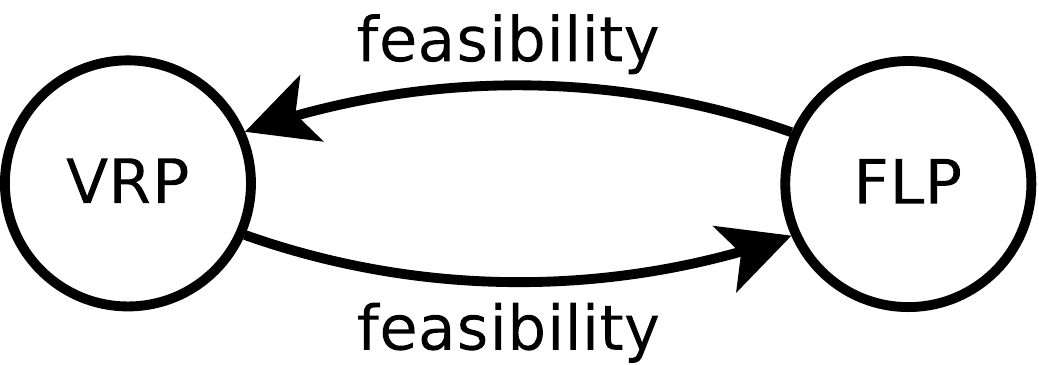}
  \caption{The LRP graph of dependency}\label{fig:lrp-dep-graph}
\end{figure}

Figure~\ref{fig:lrp-dep-graph} illustrates the graph of dependency for the Location-routing problem introduced in Chapter~\ref{chap:1}. Note that changing the FLP solution might impact both the fitness and the feasibility of solution. However, the fitness-dependency is ignored as previously explained.

\begin{align}\label{eq:overall-objective}
Z^P \colon \mathcal{\chi}^{P_1}(s^{P_2}, \dots, s^{P_n}) \times \dots \times \mathcal{\chi}^{P_n}(s^{P_1}, \dots, s^{P_{n-1}}) &\to \mathbb{R}\\
(s^{P_1}, \dots, s^{P_n}) &\mapsto z\nonumber
\end{align}

A convenient way to define $Z^P$ is by using a weighted sum of the component objectives as expressed in Equation~\ref{eq:objective-wsum}.
\begin{align}\label{eq:objective-wsum}
Z^P({s}^{P_1}, \dots, {s}^{P_n}) = \sum_{k=1}^{n} \alpha_k  Z^{P_k}(s^{P_k})
\end{align}
where $\alpha_k$ is a real number.

\begin{definition}\label{def:decision-problem}
Let $P$ be a minimisation problem and $Z^P(.)$ its objective function.
We define $\widetilde{P}$ as the decision problem associated with a given problem $P$. That is, given $k\in\mathbb{R}$, is there a solution $s$ to $\widetilde{P}$ such that $Z^P(s) \leq k$?
\end{definition}

By considering the weighted sum function, we can easilly prove --without considering of the existence of internal dependencies-- the $\mathcal{NP}$-completeness of the overall problem when all the components are in $\mathcal{NP}$ and at least one component is $\mathcal{NP}$-hard.

% (\ref{clm:p-npc})
\begin{claim}\label{clm:p-npc}
Let $P=\{P_1, \dots, P_n\}$ be a minimisation problem with $n$ interdependent components defined by the weighted sum objective function in Equation~\ref{eq:objective-wsum}. 
If $\forall P_i \in P$ such that $\widetilde{P_i}$ is in $\mathcal{NP}$ and $\exists P_j \in P$ such that $\widetilde{P_j}$ is $\mathcal{NP}$-hard, Then $\widetilde{P}$ is $\mathcal{NP}$-complete.
\end{claim}

% \begin{claimproof}
First, we need to prove that $P$ is in $\mathcal{NP}$. Since all the components of $P$ are in $\mathcal{NP}$, the solutions can be verified in polynomial time. Thus, the objective function can also be calculated in polynomial time, i.e. the overall solution $({s}^{P_1}, \dots, {s}^{P_n})$ can be verified in polynomial time.

Second, to prove that $\widetilde{P}$ is $\mathcal{NP}$-hard, we need to reduce a known $\mathcal{NP}$-hard problem to $P$ in polynomial time.
To do so, we can reduce $\widetilde{P_i}$ to $\widetilde{P}$ by setting $\alpha_i=1$, and $\alpha_j=0, \forall j \neq i$, which concludes the proof.
% \end{claimproof}

In this thesis, we are interested in problems where each component is associated to an $\mathcal{NP}$-complete decision problem.
In state-of-the-art, these problems are referred to as multi-hard problems~\cite{przybylek2016multi}, multi-silo problems~\cite{ibrahimov2012evolutionary1,ibrahimov2012evolutionary2}, or simply multi-component problems~\cite{bonyadi2013ttp}. These referred works argue that many real-world optimisation problems are composed of multiple interacting sub-problems, where each sub-problem is practically intractable.

\section{Time dependency}

Time dependency can be seen as another form of dependency between components where the notion of \textit{time window} is considered. In order to define this notion, we need to consider the following additional points.
\begin{itemize}
	\item There is a subset of components in $P$ that receive a stream of data as input. In addition, this process of data feeding can be done on multiple time slots.
	\item Some components in $P$ might have a time constraint. Which means that such components have a limited amount of time to process the data stream and generate a solution to the problem instance. If such solution cannot be generated, a smaller instance is derived such that it is possible to generate solutions for it with respect to the allowed time window. In other words, the components themselves can be decomposed into smaller sub-problems to satisfy the time constraint.
\end{itemize}

\begin{definition}[Time dependency --- high level]\label{def:time-dependency}
Let $P_i$ and $P_j$ be two components in $P$.
We say that $P_i$ is time dependent on $P_j$ (notation $P_i \xleftarrow{time}{} P_j$) if
\begin{itemize}
\item $P_i$ receives a data stream from $P_j$. This data stream depends exclusively on the instance considered for $P_j$.
\item A change of the instance considered for $P_j$ induces a change of the instance for $P_i$. More precisely, the dimension of the $P_i$ instance is dependent on the dimension of the $P_j$ instance from which $P_i$ receives the data stream.
\end{itemize}
\end{definition}

\begin{figure}[h]
  \centering
  \includegraphics[scale=0.5]{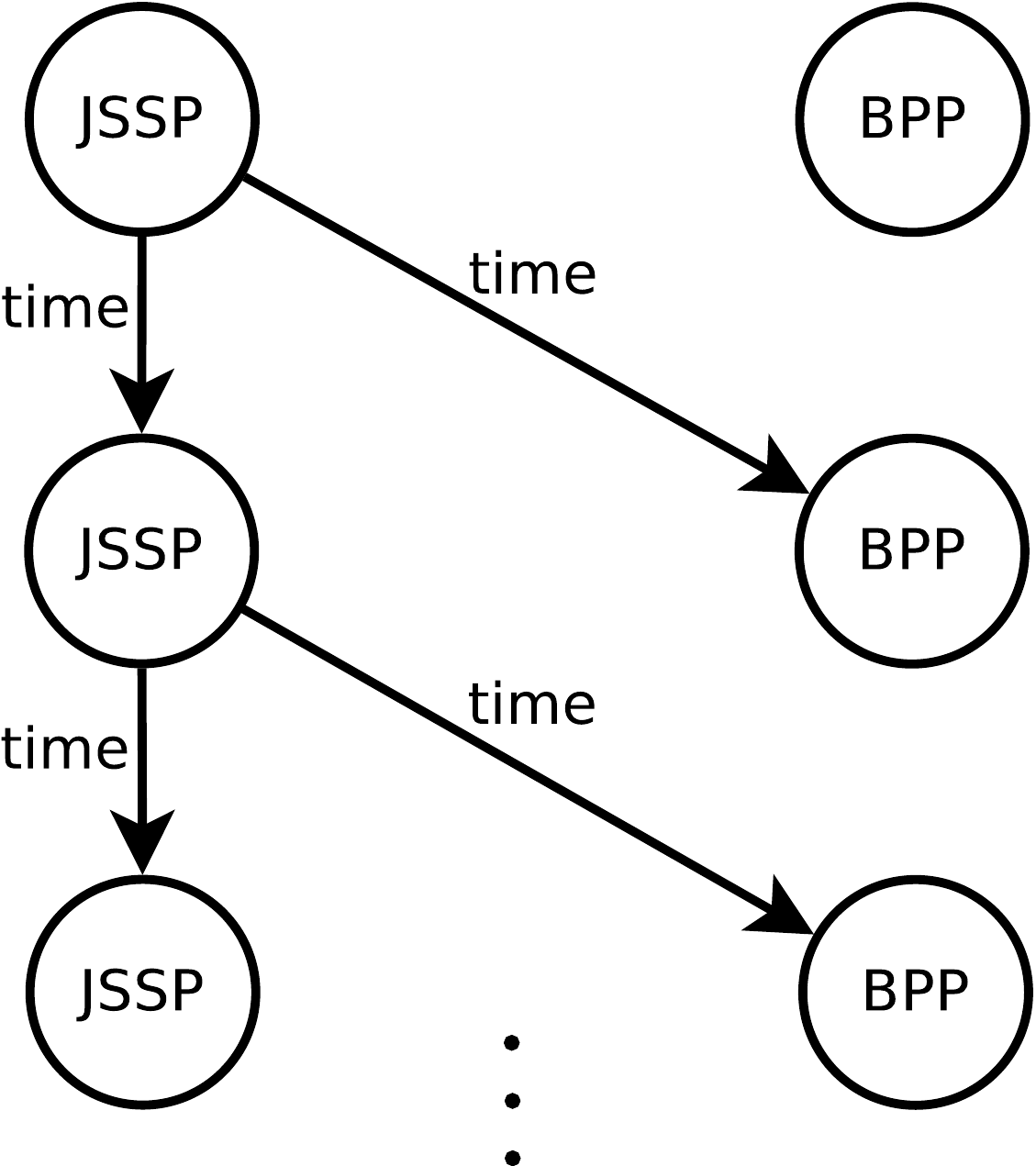}
  \caption{The graph of dependency for the problem of demand scheduling and truck loading}\label{fig:dstl-dep-graph}
\end{figure}

\begin{figure}[h]
  \centering
  \includegraphics[scale=0.5]{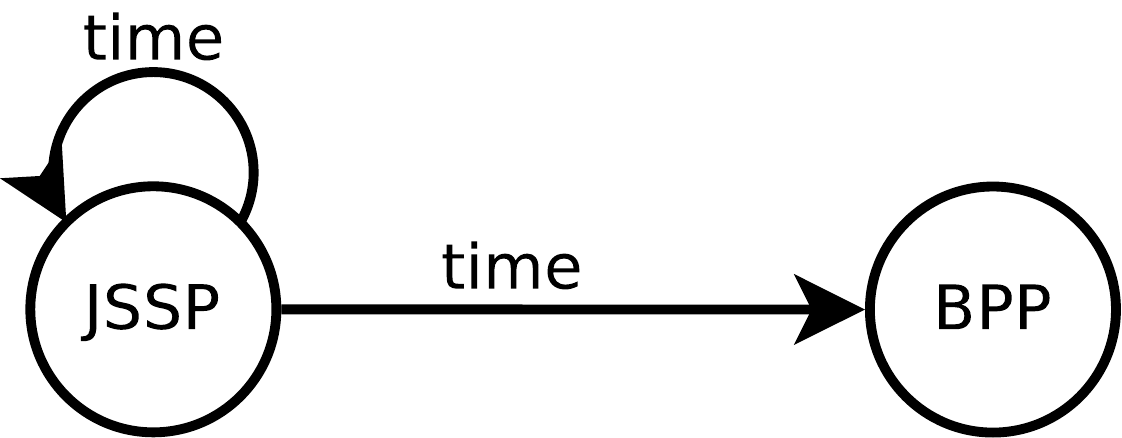}
  \caption{A compressed version of the graph of dependency for the problem of demand scheduling and truck loading}\label{fig:dstl-dep-graph-compressed}
\end{figure}

A problem with time dependencies can be modelled using a directed graph with an undetermined number of vertices, where the vertices represent the components' sub-problems, and edges represent the connections from a time window $d$ to the next one $d+1$. Figure~\ref{fig:dstl-dep-graph} illustrates the graph of dependecy for the problem of demand scheduling and truck loading briefly presented in Chapter~\ref{chap:1}.

Another possibility to model a problem with time dependencies is to use a regular graph, where the vertices represent the components and the edges represent the connections between the components' sub-problems from a time window $d$ to $d+1$ as shown in Figure~\ref{fig:dstl-dep-graph-compressed} for the same problem.

\section{Conclusion}

In this chapter, we introduced a formal model that defines the concept of dependencies in multi-component problems in a general fashion. The model is based on former research~\citet{bonyadi2016evolutionary} and our experience with multi-component problems with dependencies. 

Our goal was to propose a general mathematical model in order to distinguish and classify dependencies in multi-component optimisation problem. The model allowed us to generate the graph of dependencies of a given problem based on our definitions of multiple dependency types. We believe that these results will be important and could be used as building blocs for future theoretical investigation of multi-component problems with dependencies.

\bibliographystyle{apa}
\bibliography{references}

\end{document}